\documentclass[runningheads]{llncs}
\usepackage{graphicx}
\usepackage{indentfirst}
\usepackage{multirow}
\usepackage{color}

\begin{document}
\title{Attention Guided Network for Salient Object Detection in Optical Remote Sensing Images\thanks{This work is supported in part by the Fundamental Research
Funds for the Central Universities of China under Grant
NZ2019009.}}
\titlerunning{Attention Guided Network for SOD in Optical RSIs}
% If the paper title is too long for the running head, you can set
% an abbreviated paper title here
%    \orcidID{0000-1111-2222-3333}
\author{Yuhan Lin\inst{1} \and
Han Sun\inst{1} \and
Ningzhong Liu\inst{1} \and
Yetong Bian\inst{1} \and
Jun Cen\inst{1} \and
Huiyu Zhou\inst{2}}
\authorrunning{Yuhan Lin et al.}
% First names are abbreviated in the running head.
% If there are more than two authors, 'et al.' is used.
%
\institute{College of Computer Science and Technology, Nanjing University of Aeronautics and Astronautics, 
Nanjing, China
\\
\email{Email:sunhan@nuaa.edu.cn}
\and
School of Computing and Mathematical Sciences, University of Leicester, Leicester LE1 7RH, UK}

\maketitle              % typeset the header of the contribution
\begin{abstract}
    Due to the extreme complexity of scale and shape as well as the 
    uncertainty of the predicted location, salient object detection in optical remote sensing images (RSI-SOD) 
    is a very difficult task. The existing SOD methods can satisfy the detection performance for natural scene images, 
    but they are not well adapted to RSI-SOD due to the above-mentioned image characteristics in remote sensing images. 
    In this paper, we propose a novel Attention Guided Network (AGNet) for SOD in optical RSIs, including position enhancement 
    stage and detail refinement stage. Specifically, the position enhancement stage consists of a semantic attention module and 
    a contextual attention module to accurately describe the approximate location of salient objects. The detail refinement stage 
    uses the proposed self-refinement module to progressively refine the predicted results under the guidance of attention and 
    reverse attention. In addition, the hybrid loss is applied to supervise the training of the network, which can improve the 
    performance of the model from three perspectives of pixel, region and statistics. Extensive experiments on two popular benchmarks 
    demonstrate that AGNet achieves competitive performance compared to other state-of-the-art methods. The code will be available at
    https://github.com/NuaaYH/AGNet.

\end{abstract}
\section{Introduction}
Salient object detection (SOD) is used to simulate the human visual attention mechanism, which helps machine automatically search 
for the most attractive areas or objects in an image.

In the early days, the traditional methods for SOD are based on hand-crafted features. Such methods can meet the performance requirements 
of SOD tasks in specific scenarios, which also have the advantages of fast speed and convenient deployment. However, when application environment 
expands to a wide variety of scenarios, traditional methods do not have satisfactory generalization. In 2015, the SOD model was combined with deep 
learning technique and convolutional neural network (CNN) for the first time, which aroused the interest of researchers in the topic of CNN-based SOD methods.

At the present time, there has been a lot of excellent work on SOD tasks in natural scene images (NSIs) such as GateNet~\cite{zhao2020suppress}, F3Net~\cite{wei2020f3net}, LDF~\cite{wei2020label}, GCPANet~\cite{chen2020global}. 
However, as an emerging topic of sailency detection, the researches on SOD in optical remote sensing images (RSIs) is still relatively rare. Compared with NSIs, 
remote sensing images bring us more challenges. Specifically, RSIs are usually taken from a high altitude with a bird's-eye view through equipments such as aerial 
cameras or satellites, thus the location of salient objects may appear in various places in the frame. Meanwhile, various scales, complex shapes and indistinct 
boundaries are the unique features of objects in RSIs, which also hinder the accurate prediction of object details. Since the RSI-SOD datasets were constructed and opened, 
there have been some promising attempts to solve SOD tasks on remote sensing images, e.g, DAFNet~\cite{zhang2020dense}, EMFINet~\cite{zhou2021edge}, SARNet~\cite{huang2021semantic}. 
But none of them can solve the above-mentioned problems well, which includes the optimization of object positioning and object details.

As another huge vision task topic, attention mechanism and salient object detection have many similarities. Over the years, the attention mechanism has been widely used in 
SOD models and proved to be an extremely effective solution. In this paper, we aim at designing an attention guided network for accurate salient object detection in optical RSIs, 
called AGNet. To obtain accurate location prediction for salient objects, we propose a position enhancement stage including a semantic attention module and a contextual attention 
module, which provide the network with attention-guiding information from global semantic and local context perspectives, respectively. In order to obtain a more refined and complete 
saliency map, a self-refinement module is designed to build the detail refinement stage. The self-refinement module employs attention and reverse attention mechanisms to mine object 
details in high-confidence and low-confidence regions, thereby gradually improving the prediction results. At last, we evaluate the effectiveness of the proposed network.
\section{Related Work}
\subsection{Attention Mechanism for SOD}
In recent years, the attention mechanism is utilized to promote the performance of SOD models to a higher level. Zhao et al.~\cite{zhao2020suppress} proposes a simple gate network to filter the noise in the encoder feature map and consider the difference in contributions from different encoder blocks.
Wu et al.~\cite{wu2019cascaded} designs a holistic attention module to expand the coverage of the initial saliency map, which helps to segment the entire salient object and optimize the boundaries. Yang et al.~\cite{yang2021progressive} designs a 
branch-wise attention module to adaptively aggregate multi-scale features so that salient objects can be efficiently localized and detected. Tang et al.~\cite{tang2020class} proposes a novel non-local 
cross-level attention, which can capture long-range feature dependencies to enhance the discriminative ability of complete salient objects. And Wang et al.~\cite{wang2019salient} presents a pyramid attention 
structure to enable the network to pay more attention to saliency regions while exploiting multi-scale saliency information.
\subsection{Salient Object Detection for RSIs}
As we know, the RSI-SOD methods based on deep learning have become the main research idea since 2019. Zhang et al.~\cite{zhang2020dense} proposes a dense attention flow network, in which attention information flows from shallow layers 
to deep layers to guide the generation of high-level feature maps. Cong et al.~\cite{cong2021rrnet} presents a graph structure-based reasoning module and a parallel multi-scale attention module, which greatly improve the accuracy and 
completeness of the saliency map. Zhou et al.~\cite{zhou2021edge} uses three encoders with different input scales to extract and fuse multi-scale features, and then introduces the guidance of edge information to obtain high-quality 
saliency maps. Huang et al.~\cite{huang2021semantic} integrates multiple high-level features to locate the object position and feed back the high-level features into the stage of shallow feature fusion. And in~\cite{tu2021orsi}, a joint learning scheme based on bidirectional feature transformation
is proposed to simultaneously optimize the boundaries and regions of salient objects.

In general, since the above deep learning-based RSI-SOD methods have made great progress, these solutions still show unsatisfactory performance when dealing with some difficult samples in Fig.~\ref{sota}. 
In our AGNet, we design an intuitively more natural method according to the image features of RSIs, which is implemented by position enhancement stage and detail refinement stage.
\section{Approach}
\begin{figure}[!ht]
    \centering
    \includegraphics[width=4in]{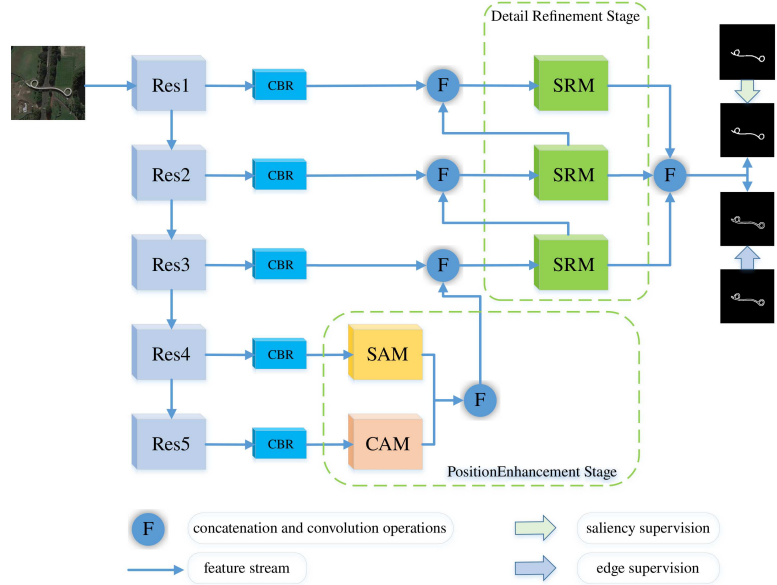}
    \caption{The architecture of AGNet. CBR refers to the combination of convolutional layer, batch nomalization operation, and activation function. SAM means semantic attention module, 
    CAM denotes contextual attention module, and SRM refers to self-refinement module.} \label{OR}
\end{figure}

In this section, we will specify the details of the proposed network. The overall framework of AGNet is given in Fig.~\ref{OR}. From the figure, it is clear that the inference process of the proposed network 
consists of three main parts: feature encoding backbone, position enhancement stage and detail refinement stage. In the following, the above three components will be discussed in turn. Finally, 
we will also present the design of the used hybrid loss.
\subsection{Feature Encoding Backbone}
In this paper, we adopt the widely used Res2Net-50~\cite{gao2019res2net} as the feature encoding backbone of AGNet.
%In many vision tasks, high-performing classification networks are often used as the backbone for feature encoding. In this paper, we extract five sets of feature maps of different scales using 
%a Res2Net-50 network pre-trained on the ImageNet dataset. Res2Net, proposed by Cheng et al.~\cite{gao2019res2net} is an improvement of the classical classification network ResNet~\cite{he2016deep}, which constructs hierarchical 
%residual connections within a single residual block to improve the module's ability to extract multi-scale features. In RSI-SOD, a powerful multi-scale feature extraction capability is essential for 
%detecting scale-variant salient objects, thus we choose Res2Net-50 to undertake this part of the work.
To reduce the computational complexity and the overall number of parameters in our network, a ${1 \times 1}$ convolutional layer is used to perform channel downscaling on the five convolutional block outputs of 
Res2Net. At last, all the five groups of feature maps are converted to 128 channels and ready for the next processing step. This process can be formulated as:
\begin{equation}
F_i  = relu(bn(conv_{1 \times 1} (f_i ))),i = 1...5
\end{equation}
in which $conv_{1 \times 1}$ denotes the ${1 \times 1}$ convolution operation, $bn$ refers to batch normalization operation, and $relu$ denotes the ReLU activation function. These series 
of operations will be represented by $CBR$ in this paper.
\subsection{Position Enhancement Stage}
Inspired by the coarse-to-fine strategy, we propose a position enhancement stage to locate potentially salient object position in the deep feature map. Specifically, the semantic attention module (SAM) is proposed to model semantic 
relations between different channels on the fourth layer feature map and the contextual attention module (CAM) is designed to estimate the accurate location information of salient objects on the last layer.
\subsubsection{Semantic Attention Module.} 
\begin{figure}[!ht]
    \centering
    \includegraphics[width=4in]{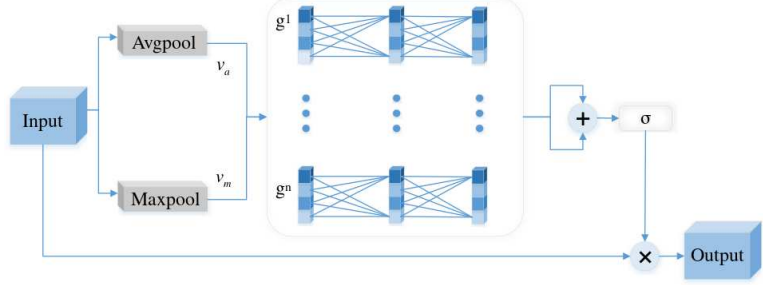}
    \caption{The architecture of SAM.} \label{SAM}
\end{figure}

SENet~\cite{hu2018squeeze} is the most widely used method for calculating channel attention. However, we think that the dimensionality reduction of channels in SENet will have the side effect of making some semantic information 
lost. Also, capturing the dependencies between all channels is inefficient and unnecessary~\cite{wang2020efficient}. Therefore, a novel semantic attention module without dimensionality reduction is designed in Fig.~\ref{SAM}, which can be efficiently 
implemented via one-dimensional group convolution.

Considering that the fourth layer feature map already has a large enough receptive field to describe the semantic information, we use this feature map as input to SAM. First, 
global average pooling and global maximum pooling are used on the feature map to obtain a global semantic representation of the features, denoted as $v_a$ and $v_m$. Then, we design a unique method for computing channel 
attention based on group convolution. Specifically, the above two channel vectors are used to compute local semantic correlations by two one-dimensional group convolutional layers. Each one-dimensional convolution 
kernel is computed with adjacent channel features, and thus the convolution results can reflect local cross-channel semantic interactions. In this paper, SAM is able to provide the maximum contribution to the network 
when we set the number of groups to 8. The semantic attention can be formulated as:
\begin{equation}
att = \sigma (gconv_{1 \times 1} (avgpool(f_4 )) + gconv_{1 \times 1} (maxpool(f_4 )))
\end{equation}
where $\sigma$ denotes the sigmoid activation function, $gconv_{1 \times 1}$ denotes the two ${1 \times 1}$ group convolutional layers, $avepool$ and $maxpool$ are the average-pooling and max-pooling in spatial dimension, respectively.
\subsubsection{Contextual Attention Module.}In RSIs, the location of the salient object is more randomly distributed than in NSIs, such as the center of the view or different corners of the image. Therefore, the contextual attention module is proposed to guide the network to provide the most accurate object position in the deepest layer.

\begin{figure}[!ht]
    \centering
    \includegraphics[width=4in]{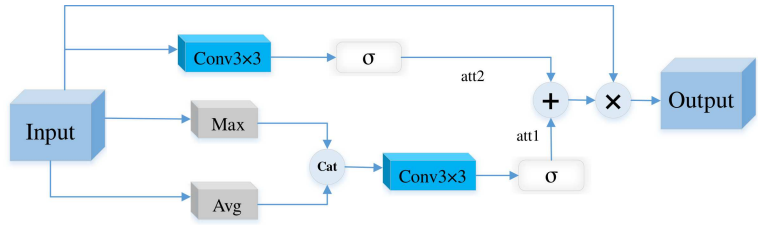}
    \caption{The architecture of CAM.} \label{CAM}
\end{figure}

As shown in Fig.~\ref{CAM}, unlike previous spatial attention, our CAM calculates the saliency of the location from the channel dimension and the local spatial dimension, respectively. Specifically, one branch performs maximum pooling 
and average pooling on the input in the channel dimension, and then cascades the two single-channel feature maps to generate a spatial attention map by convolutional fusion. The other branch computes an attention map directly on the input using ${3 \times 3}$ convolution, 
which reflects the influence of neighboring pixels on the saliency of a location in the local spatial dimension. The contextual attention can be formulated as:
\begin{equation}
    att1 = \sigma (conv_{3 \times 3} (cat(max (f_5 ),avg(f_5 ))))
\end{equation}
\begin{equation}
    att2 = \sigma (conv_{3 \times 3} (f_5 ))
\end{equation}
in which $avg$ and $max$ are the average-pooling and max-pooling in channel dimension, $cat$ denotes
feature concatenation along channel axis.
\subsection{Detail Refinement Stage}
\begin{figure}[!ht]
    \centering
    \includegraphics[width=4.2in]{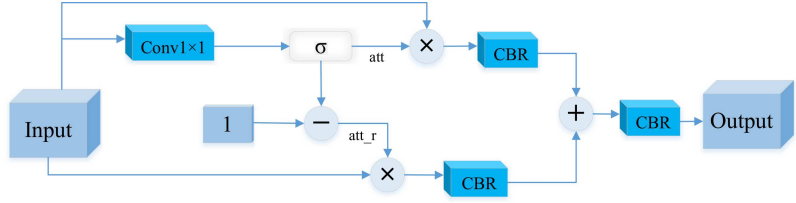}
    \caption{The architecture of SRM.} \label{SRM}
\end{figure}

As shown in most researches, the shallow features of the network often tend to provide some additional information in terms of local details to the final prediction results. In this section, we will introduce the most critical component of the 
detail refinement stage, the self-refinement module (SRM), as shown in Fig.~\ref{SRM}.

Specifically, the input to each SRM are the fused feature map, which are obtained from the deep feature of the previous layer and shallow feature of the corresponding layer. 
Subsequently, the convolution operation is used to compute the attention map reflecting the salient regions. At the same time, we invert this attention map to obtain the reverse attention map, which reflects some low-confidence regions ignored by 
the network. These two attention maps are multiplied with the fused feature map to guide the network to further mine the saliency information from the high-confidence salient regions and the low-confidence background regions. Finally, the two results are 
summed and fused by the convolutional layer to obtain the output of the SRM. This process can be formulated as:
\begin{equation}
att = \sigma (conv_{1 \times 1} (f)),att\_r = 1 - att
\end{equation}
\begin{equation}
F = cbr(cbr(att*f) + cbr(att\_r*f))
\end{equation}

At last, to fully utilize the effect of SRM, the results of the three SRM modules are integrated to obtain the final prediction map of the network by concatenation and convolution operations, as shown in Fig.~\ref{OR}.
\subsection{Hybrid Loss of AGNet}
In this paper, the widely used cross-entropy loss, IoU loss and recent F-value loss are introduced into hybrid loss to supervise the training of the network from the perspectives of pixels, regions, statistics.
Furthermore, edge information is also added to the loss function. 
In the end, the total loss of the network can be described by the following equation:
\begin{equation}
loss = loss_{bce}  + \lambda  \cdot loss_{iou}  + loss_f 
\end{equation}
\begin{equation}
Loss = loss(P,G_p ) + \mu  \cdot loss(E,G_e )
\end{equation}
where $\lambda$ and $\mu$ are hyperparameters that balance the contributions of the three losses and the contributions of the two supervised objects. Empirically, we set $\lambda  = 0.6$ and $\mu  = 0.5$. 
\section{Experiment}
%\subsection{Datasets and Evaluation Metrics}
%The widely used salient object detection datasets in remote sensing image are ORSSD~\cite{li2019nested} and EORSSD~\cite{zhang2020dense}. The EORSSD dataset is an extended version of the ORSSD dataset, which is expanded from 800 images to 2000 images. In order to facilitate 
%training and evaluation, the creators of the dataset divide ORSSD into a training set of 600 images and a test set of 200 images, and divide EORSSD into a training set of 1400 images and a test set of 600 images. 

%For a fair comparison with other methods, we use four commonly used evaluation metrics to evaluate our model performance. They are the mean absolute error (MAE),
%mean F-measure (mF), S-measure (Sm) and mean E-measure (mE).
\subsection{Implementation Details}
For the EORSSD~\cite{zhang2020dense} dataset and ORSSD~\cite{li2019nested} dataset, we expand their training sets to 11200 and 4800 samples by random flipping and random rotating. We train the proposed AGNet using the adam optimizer with initial learning rate of 1e-4, weight decay of 5e-4, and batch size of 8. Cosine annealing 
decay strategy is adopted to adjust the learning rate and the minimum learning rate is set to 1e-5. Our network is totally trained for 60 epochs and the input size is $224 \times 224$.
For a fair comparison with other methods, four commonly used metrics including the mean absolute error (MAE),
mean F-measure (mF), S-measure (Sm) and mean E-measure (mE) are adopted to evaluate the model performance as suggested in~\cite{huang2021semantic}.
\begin{figure}[!ht]
    \centering
    \includegraphics[width=3in]{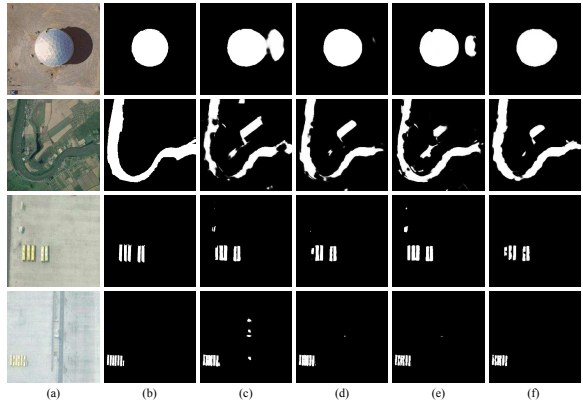}
    \caption{Comparison of predictions among different networks. (a) Image; (b) Ground truth; (c) Baseline; (d) Baseline+PES; (e) Baseline+DRS; (f) Baseline+PES+DRS.} \label{ab}
\end{figure}
\subsection{Ablation Studies}
To demonstrate the effectiveness of the proposed method with convincing data, a series of ablation experiments are presented in this section. The same experimental parameter settings are used throughout the experiment. The qualitative comparison and quantitative comparison are shown in Fig.~\ref{ab} 
and Table~\ref{tab1}. Specifically, Baseline denotes the U-shaped model with Res2Net-50 as the backbone, PES means the position enhancement stage and DRS refers to the detail refinement stage.
\subsubsection{Effectiveness of PES.}From Table~\ref{tab1}, it can be found that the addition of PES makes the performance of the network have obvious improvements in three main metrics, such as a 
gain of 0.83\% in mF. The effectiveness of PES can be more thoroughly demonstrated in Fig. 5. The fourth column of each row is the saliency map predicted by the model combined with PES on the basis of the baseline. For example, the pictures in the first and second rows in Fig.~\ref{ab} are representative of 
challenging scenes in the EORSSD dataset. The salient object in the image of the first row is accompanied by interference information such as shadows, while in the image of the second row, the color and texture features of the river are extremely similar to its surroundings, which can easily guide the 
network to output wrong results. Thanks to our proposed PES, the model has the ability to remove the background interference and accurately locate the object area, as shown in the third and fourth columns in Fig.~\ref{ab}.  
\subsubsection{Effectiveness of DRS.}To prove the effectiveness of our DRS, we additionally train with a combination of baseline and DRS. The trained model shows the contribution of DRS to the network in Table~\ref{tab1}, such as a 0.5\% improvement on mF. The third and fourth rows of images in Fig.~\ref{ab} are 
used to illustrate the effectiveness of DRS. As can be seen from the fifth column of Fig.~\ref{ab}, the gaps between closely spaced vehicles are more apparent compared to the baseline, while the shapes of the vehicles are more fully highlighted. This change is because DRS improves the prediction quality of 
the object details by the network. Our method can refine and supplement the details of salient objects, which is helpful for the application of RSI-SOD in earth observation, object counting, etc.
\begin{table}[!ht]
    \caption{Quantitative evaluation of ablation studies on the EORSSD dataset.}
    \label{tab1}
    \centering
    \setlength{\tabcolsep}{3mm}{
    \begin{tabular}{|l|l|l|l|}
    \hline
        \multirow{2}{*}{Methods} & \multicolumn{3}{c|}{EORSSD} \\ \cline{2-4} 
        ~ & mF$\uparrow$ & mE$\uparrow$ & Sm$\uparrow$  \\ \hline
        Baseline & 0.8613 & 0.9554 & 0.9229  \\ %\hline
        Baseline+PES & 0.8696 & 0.9599 & 0.9248  \\ %\hline
        Baseline+DRS & 0.8663 & 0.9578 & 0.9247  \\ %\hline
        Baseline+PES+DRS & 0.8736 & 0.9614 & 0.9284  \\ \hline
    \end{tabular}}
\end{table}
\subsection{Comparison with State-of-the-arts}
\begin{figure}[!ht]
    \centering
    \includegraphics[width=4.1in]{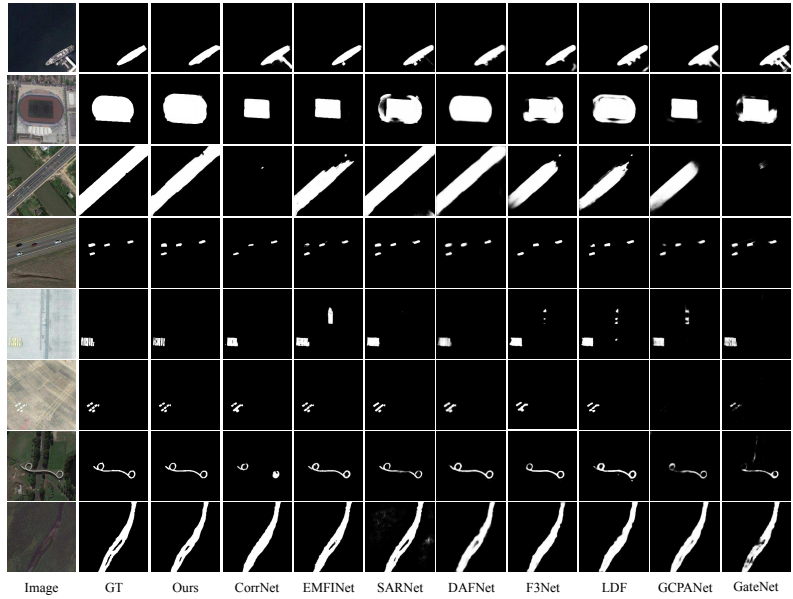}
    \caption{Visual comparisons of different methods} \label{sota}
\end{figure}

To demonstrate the validity of our proposed network, we compare the proposed method against other fourteen current state-of-the-art (SOTA) methods, including eight SOD methods for NSIs (i.e., ITSD~\cite{zhou2020interactive}, LDF~\cite{wei2020label}, MINet~\cite{pang2020multi}, GCPANet~\cite{chen2020global}, GateNet~\cite{zhao2020suppress}, F3Net~\cite{wei2020f3net},  
PA-KRN~\cite{xu2021locate} and SUCA~\cite{li2020stacked}), and six recent SOD methods for optical RSIs (i.e., CorrNet~\cite{li2022lightweight}, DAFNet~\cite{zhang2020dense}, EMFINet~\cite{zhou2021edge}, LVNet~\cite{li2019nested}, MJRBM~\cite{tu2021orsi} and SARNet~\cite{huang2021semantic}). All NSI-SOD methods are retrained using the original code projects published by their authors. 
For RSI-SOD methods, we use the saliency maps provided on the corresponding methods on github. Table~\ref{tab2} and Fig.~\ref{sota} show the comparison of the different methods on the four metrics and visualization results, respectively.
\begin{table}[!ht]
    \caption{Quantitative evaluation. The best three results are highlighted in red, blue, green.}
    \label{tab2}
    \centering
    \setlength{\tabcolsep}{1mm}{
    \begin{tabular}{|l|l|l|l|l|l|l|l|l|l|}
    \hline
        \multirow{2}{*}{Methods} & \multirow{2}{*}{type} & \multicolumn{4}{c|}{EORSSD} & \multicolumn{4}{c|}{ORSSD} \\ \cline{3-10}
        ~ & ~ & mF$\uparrow$ & MAE$\downarrow$ & Sm$\uparrow$ & mE$\uparrow$ & mF$\uparrow$ & MAE$\downarrow$ & Sm$\uparrow$ & mE$\uparrow$  \\ \hline
        ITSD & NSI & 0.8313 & 0.0109 & 0.9043 & 0.9265 & 0.8590 & 0.0172 & 0.9056 & 0.9410  \\ %\hline
        LDF & NSI & 0.8317 & 0.0087 & 0.9078 & 0.9464 & 0.8779 & 0.0137 & 0.9143 & 0.9518  \\ %\hline
        MINet & NSI & 0.8233 & 0.0097 & 0.9049 & 0.9190 & 0.8622 & 0.0153 & 0.9040 & 0.9355  \\ %\hline
        GCPANet & NSI & 0.7968 & 0.0104 & 0.8855 & 0.8993 & 0.8488 & 0.0177 & 0.9005 & 0.9253  \\ %\hline
        GateNet & NSI & 0.8314 & 0.0099 & 0.9090 & 0.9238 & 0.8769 & 0.0144 & 0.9192 & 0.9472  \\ %\hline
        PA-KRN & NSI & 0.8434 & 0.0109 & 0.9194 & 0.9416 & 0.8824 & 0.0145 & 0.9243 & 0.9566  \\ %\hline
        F3Net &   NSI & 0.8164 & 0.0088 & 0.9030 & 0.9401 & 0.8677 & 0.0155 & 0.9160 & 0.9538  \\ %\hline
        %EGNet & NSI & 0.7012 & 0.0113 & 0.8593 & 0.8574 & 0.7587 & 0.0219 & 0.8725 & 0.8937  \\ \hline
        %DSS & NSI & 0.5933 & 0.0188 & 0.7868 & 0.7589 & 0.7081 & 0.0366 & 0.8261 & 0.8335  \\ \hline
        %PoolNet & NSI & 0.6416 & 0.0208 & 0.8214 & 0.8000 & 0.7090 & 0.0360 & 0.8396 & 0.8584  \\ \hline
        %R3Net & NSI & 0.6479 & 0.0174 & 0.8192 & 0.8262 & 0.7457 & 0.0403 & 0.8171 & 0.8609  \\ \hline
        %RADF & NSI & 0.6809 & 0.0167 & 0.8211 & 0.8555 & 0.6981 & 0.0384 & 0.8256 & 0.8274  \\ \hline
        SUCA & NSI & 0.8040 & 0.0102 & 0.8971 & 0.9131 & 0.8347 & 0.0150 & 0.8992 & 0.9349  \\ %\hline
        CorrNet & RSI & {\color{blue}0.8690} & 0.0087 & {\color{blue}0.9297} & {\color{blue}0.9512} & {\color{green}0.9077} & 0.0107 & {\color{blue}0.9405} & {\color{blue}0.9687}  \\ %\hline
        DAFNet & RSI & 0.8135 & {\color{red}0.0060} & 0.9175 & 0.9295 & 0.8578 & 0.0111 & 0.9186 & 0.9496  \\ %\hline
        EMFINet & RSI & 0.8601 & {\color{green}0.0079} & {\color{red}0.9307} & 0.9445 & {\color{blue}0.9083} & {\color{blue}0.0104} & {\color{red}0.9435} & {\color{green}0.9662}  \\ %\hline
        LVNet & RSI & 0.7488 & 0.0146 & 0.8639 & 0.8729 & 0.8107 & 0.0211 & 0.8807 & 0.9204  \\ %\hline
        MJRBM & RSI & 0.8162 & 0.0105 & 0.9068 & 0.9087 & 0.8679 & 0.0151 & 0.9190 & 0.9338  \\ %\hline
        SARNet & RSI & {\color{green}0.8625} & 0.0091 & {\color{green}0.9284} & {\color{green}0.9482} & 0.8942 & {\color{green}0.0105} & 0.9361 & 0.9596  \\ %\hline
        AGNet & RSI & {\color{red}0.8736} & {\color{blue}0.0069} & {\color{green}0.9284} & {\color{red}0.9614} & {\color{red}0.9109} & {\color{red}0.0093} & {\color{green}0.9392} & {\color{red}0.9707}  \\ \hline
    \end{tabular}}
\end{table}
\subsubsection{Quantitative comparison.}In Table~\ref{tab2}, we report the mF, mE, Sm, MAE of our method and fourteen other methods on two RSI-SOD datasets. Compared with other RSI-SOD methods, our method shows impressive performance. Specifically, on the EORSSD dataset, AGNet is much higher than the second-ranked method in 
mF, mE, e.g., mF: 0.8736 (AGNet) v.s. 0.8690 (CorrNet), and mE: 0.9614 (AGNet) v.s. 0.9512(CorrNet). Meanwhile, on MAE, our method is only slightly weaker than DAFNet by a gap of 0.0009. While on the ORSSD dataset, our method ranks first on mF, mE and MAE. Although our method fails to achieve the best score on 
both datasets on Sm, our method outperforms by achieving more impressive performance with smaller parameters, e.g., param: 26.60M (AGNet) vs 107.26M (EMFINet). And our AGNet also beats SARNet with the same scale of parameters, the former has about 26M and the latter has 40M. 
In competition with all NSI-SOD methods, our model and most RSI-SOD methods are a lot ahead of them, which further demonstrates the 
necessity of specially designing the model for the SOD of optical RSIs. 
\subsubsection{Qualitative comparison.}In Fig.~\ref{sota}, we show the saliency maps for the more difficult test samples. From the visual analysis of 
the saliency maps in the first to third rows, AGNet outperforms other methods when dealing with multi-scale objects, which is exactly the benefit of a powerful multi-scale feature extractor and multi-layer feature fusion strategy.
The images in the fourth to sixth rows involve multiple small salient objects, which is a challenging problem in the RSI-SOD task. Our method can detect all salient objects and get refined segmentation results. For example, there are seven circular buildings in the image in the sixth row, all of which 
are clearly marked by our method with the help of PES. But other methods can only detect a part of the buildings, or make objects blend together, such as GateNet and DAFNet. In the last two rows in Fig.~\ref{sota}, we present the performance of all methods when encountering slender or complex-structured objects. For the ring-shaped buildings 
in the seventh row, our method provides the most complete and clear detection results due to the effect of DRS. For the two lands surrounded by the river in the eighth row, other methods either conflate them with the river or only show a hole in the saliency map,
while our method successfully recognizes the two lands as the background, which further confirms the importance of self-refinement for the RSI-SOD task. 
%In summary, the proposed 
%PES enables the network to accurately estimate the location of the object, thereby completely detecting the location of the car in fourth and fifth rows. And the designed DRS allows the saliency map to fully display the details of the salient object through continuous self-refinement, especially as can be seen from sixth and eighth rows.
\section{Conclusion}
In this paper, a novel attention-guided network is proposed to solve the RSI-SOD task. To enhance the detection quality of salient object position, a semantic attention module and a contextual attention module are jointly used to guide the network to detect coarse but accurate object locations. In order to make the details of the 
object more refined and complete, a self-refinement module is designed to refine the feature map from high-confidence and low-confidence regions. Finally, an efficient hybrid loss is employed in the training of the network to boost performance. Experimental results demonstrate the robustness of our method in dealing with multi-scale 
objects, multiple small objects, and objects with complex structures. In the future, we will try to design a high-precision and lightweight network structure to further promote the application of the RSI-SOD model in real life.

\iffalse 
For citations of references, we prefer the use of square brackets
and consecutive numbers. Citations using labels or the author/year
convention are also acceptable. The following bibliography provides
a sample reference list with entries for journal
articles~\cite{ref_article1}, an LNCS chapter~\cite{ref_lncs1}, a
book~\cite{ref_book1}, proceedings without editors~\cite{ref_proc1},
and a homepage~\cite{ref_url1}. Multiple citations are grouped
\cite{ref_article1,ref_lncs1,ref_book1},
\cite{ref_article1,ref_book1,ref_proc1,ref_url1}.
\fi
%
% ---- Bibliography ----
%
% BibTeX users should specify bibliography style 'splncs04'.
% References will then be sorted and formatted in the correct style.
%
\bibliographystyle{splncs04}
\bibliography{mybib}
%
\iffalse 

\fi
\end{document}